\documentclass[11pt,a4paper]{article}
\usepackage[hyperref]{emnlp-ijcnlp-2019}
\usepackage{times}
\usepackage{latexsym}
\usepackage{graphicx}
\usepackage{url}
\usepackage{hyperref}
\usepackage{amsmath}
\usepackage{amsfonts}
\usepackage{enumitem}
\usepackage{multirow}
\usepackage{hhline}
\usepackage{tikz}
\usepackage{tablefootnote}
\usepackage{todonotes}
\usepackage{booktabs}

\aclfinalcopy 

\title{Encoders Help You Disambiguate Word Senses \\in Neural Machine Translation}

\author{Gongbo Tang$^1$\quad Rico Sennrich$^{2,3}$\quad Joakim Nivre$^1$ \medskip\\
  $^1$Department of Linguistics and Philology, Uppsala University\\
  $^2$Institute of Computational Linguistics, University of Zurich\\
  $^3$School of Informatics, University of Edinburgh\\
  {\tt firstname.lastname@\{lingfil.uu.se, ed.ac.uk\}}}

\date{}

\begin{document}
\maketitle
\begin{abstract}
  Neural machine translation (NMT) has achieved new state-of-the-art performance in translating ambiguous words. However, it is still unclear which component dominates the process of disambiguation. In this paper, we explore the ability of NMT encoders and decoders to disambiguate word senses by evaluating hidden states and investigating the distributions of self-attention. We train a classifier to predict whether a translation is correct given the representation of an ambiguous noun. We find that encoder hidden states outperform word embeddings significantly which indicates that encoders adequately encode relevant information for disambiguation into hidden states. Decoders could provide further relevant information for disambiguation.  Moreover, the attention weights and attention entropy show that self-attention can detect ambiguous nouns and distribute more attention to the context. Note that this is a revised version of \newcite{tang-etal-2019-encoders}. The content related to decoder hidden states has been updated. 
\end{abstract}

\section{Introduction}

Neural machine translation (NMT) models \cite{kal2013recurrent,sutskever2014sequence,cho2014learning,bahdanau15joint,luong2015effective} have access to the whole source sentence for the prediction of each word, which intuitively allows them to perform word sense disambiguation (WSD) better than previous phrase-based methods, and \newcite{rios2018wsd} have confirmed this empirically. However, it is still unclear which component dominates the ability to disambiguate word senses. 
We explore the ability of NMT encoders and decoders to disambiguate word senses by evaluating hidden states and investigating the self-attention distributions.

\newcite{marvin2018exploring} find that the hidden states in higher encoder layers do not perform disambiguation better than those in lower layers and conclude that encoders do not encode enough relevant context for disambiguation. However, their results are based on small data sets, and we wish to revisit this question with larger-scale data sets. 
\newcite{Tang2018WSD} speculate that encoders have encoded the relevant information for WSD into hidden states before decoding but without any experimental tests. 

In this paper, we first train a classifier for WSD, on a much larger data set than \newcite{marvin2018exploring}, extracted from ContraWSD \cite{rios2017improving}, for both German$\rightarrow$English (DE$\rightarrow$EN) and German$\rightarrow$French (DE$\rightarrow$FR). The classifier is fed a representation of ambiguous nouns and a word sense (represented as the embedding of a translation candidate), and has to predict whether the two match.
We can learn the role that encoders play in encoding information relevant for WSD by comparing different representations: word embeddings and encoder hidden states at different layers. 
We extract encoder hidden states from both RNN-based (\textit{RNNS2S}) \cite{luong2015effective} and \textit{Transformer} \cite{vaswani2017Attention} models. 
\newcite{belinkov2017what,belinkov2017evaluating} have shown that the higher layers are better at learning semantics. We hypothesize that the hidden states in higher layers incorporate more relevant information for WSD than those in lower layers.
In addition to encoders, we also probe how much do decoder hidden states contribute to the WSD classification task. 

Recently, the distributions of attention mechanisms have been used for interpreting NMT models \cite{ghader2017what,voita2018context,Tang2018WSD,voita-etal-2019-analyzing,tang2019understanding}. We further investigate the attention weights and attention entropy of self-attention in encoders to explore how self-attention incorporates relevant information for WSD into hidden states. 
As sentential information is helpful in disambiguating ambiguous words, we hypothesize that self-attention pays more attention to the context when modeling ambiguous words, compared to modeling words in general. 

Here are our findings:
\begin{itemize}[noitemsep]
  \item Encoders encode lots of relevant information for word sense disambiguation into hidden states, even in the first layer. The higher the encoder layer, the more relevant information is encoded into hidden states. 
  \item Forward RNNs are better than backward RNNs in modeling ambiguous nouns. 
  \item Self-attention focuses on the ambiguous nouns themselves in the first layer and keeps extracting relevant information from the context in higher layers. 
  \item Self-attention can recognize the ambiguous nouns and distribute more attention to the context words compared to dealing with nouns in general. 
\end{itemize}

\section{Methodology} 
\label{sec:methodology}

\subsection{WSD Classifier}

\textit{ContraWSD} \citep{rios2017improving} is a WSD test set for NMT. Each ambiguous noun in a specific sentence has a small number of translation candidates. We generate instances that are labelled with one candidate and a binary value indicating whether it corresponds to the correct sense. 

\paragraph{Encoders} 
Given an input sentence, NMT encoders generate hidden states of all input tokens. Our analysis focuses on the hidden states of ambiguous nouns ($R_{ambi}$). We use word embeddings from NMT models to represent the translation candidates ($R_{sense}$). If the ambiguous nouns or translation candidates are split into subwords, we just sum the representations. 

\begin{figure}[htbp]
\centering
        \includegraphics[totalheight=4.1cm]{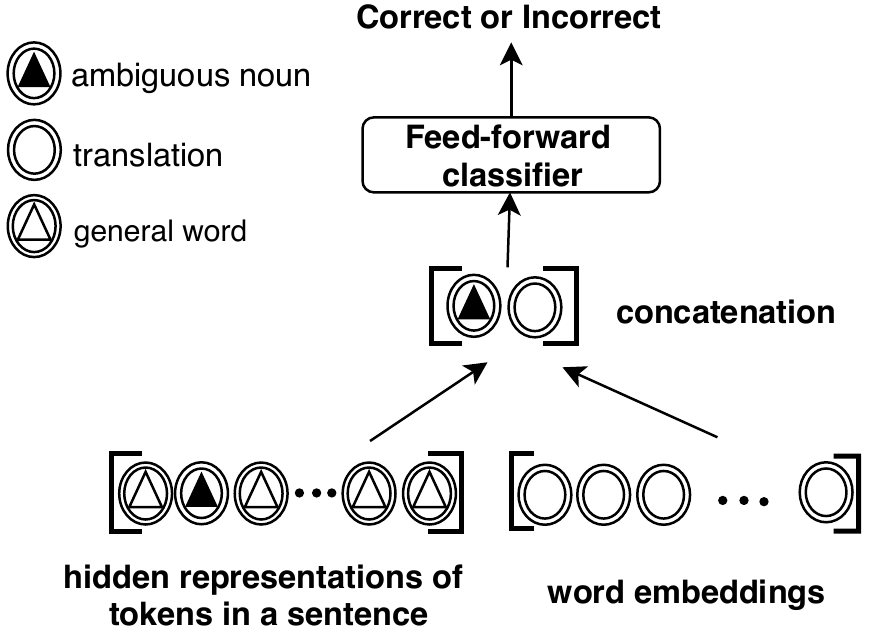}
    \caption{Illustration of the WSD classification task, using encoder hidden states to represent ambiguous nouns. The input of the classifier is the concatenation of the ambiguous word and the translation. The output of the classifier is ``correct'' or ``incorrect''. }
    \label{fig:classifier}
\end{figure}  

Figure \ref{fig:classifier} illustrates the WSD classification task. We first generate hidden states for each sentence. The classifier is a feed-forward neural network with only one hidden layer. The input of the classifier is the concatenation of $R_{ambi}$ and $R_{sense}$. The classifier predicts whether the translation is the correct sense of the ambiguous noun. 

As the baseline, we use word embeddings from NMT models as representations of ambiguous nouns. Each ambiguous noun has only one corresponding word embedding, so such a classifier can at best learn a most-frequent-sense solution, while hidden states are based on sentential information, so that ambiguous nouns have different representations in different source sentences. We can learn to what extent relevant information for WSD is encoded by encoders by comparing to the baseline. 

\paragraph{Decoders} 
To explore the role of decoders, we feed the decoder hidden state at the time step predicting the translation of the ambiguous noun, and the word embedding of the current translate candidate into the classifier. The decoder hidden state is extracted from the last decoder layer. 
To get these hidden states. we force NMT models to generate the reference translations using constrained decoding \cite{Post2018fast}. 
Since decoders are crucial in NMT, we assume that the decoder hidden states incorporate more relevant information for WSD from the decoder side. Thus, we hypothesize that using decoder hidden states can achieve better WSD performance.

\subsection{Attention Distribution}
\label{sub:attention_distribution}

The attention weights can be viewed as the degree of contribution to the current word representation, which provides a way to interpret NMT models. 
\newcite{Tang2018why} have shown that Transformers with self-attention are better at WSD than RNNs. However, the working mechanism of self-attention has not been explored. We try to use the attention distributions in different encoder layers to interpret how self-attention incorporates relevant information to disambiguate word senses. 

All the ambiguous words in the test set are nouns. \newcite{ghader2017what} have shown that nouns have different attention distributions from other word types. Thus, we compare the attention distributions of ambiguous nouns to nouns in general\footnote{We use \textit{TreeTagger} \cite{Schmid1995treetagger} to identify nouns. } in two respects. One is the attention weight over the word itself. The other one is the concentration of attention distributions. We use attention entropy \cite{ghader2017what} to measure the concentration. 
\begin{equation} \label{eq:attention-entropy}
E_{At}(x_{t}) = - \sum_{i=1}^{|x|} At(x_{i},x_{t}) \log At(x_{i},x_{t}) 
\end{equation} 
Here $x_{i}$ denotes the $i$th source token, $x_{t}$ is the current source token, and $At(x_{i},x_{t})$ represents the attention weight from $x_{t}$ to $x_{i}$. We merge subwords after encoding, following the method in \newcite{koehn2017challenges}.\footnote{(1) If a query word is split into subwords, we add their attention weights. (2) If a key word is split into subwords, we average their attention weights.} Each self-attention layer has multiple heads and we average the attention weights from all the heads. 

In theory, sentential information is more important for ambiguous words that need to be disambiguated than non-ambiguous words. From the perspective of attention weights, for ambiguous words, we hypothesize that self-attention distributes more attention to the context words to capture the relevant sentential information, compared to words in general. 
From the perspective of attention entropy, we hypothesize that self-attention focuses on the related context words rather than the entire sentence which produces a smaller entropy. 
If ambiguous words have a lower weight and a smaller entropy than words in general, the results can confirm our hypotheses. 

\section{Experiments}

For NMT models, we use the \textit{Sockeye} \cite{Hieber2017sockeye} toolkit to train \textit{RNNS2S}s and \textit{Transformer}s. DE$\rightarrow$EN training data is from the WMT17 shared task \cite{wmt17}. DE$\rightarrow$FR training data is from Europarl (v7) \cite{koehn2005epc} and News Commentary (v11) cleaned by \newcite{rios2017improving}.\footnote{\url{http://data.statmt.org/ContraWSD/}} 
In \textit{ContraWSD}, each ambiguous noun has a small number of translation candidates. The average number of word senses per noun is 2.4 and 2.3 in DE$\rightarrow$EN and DE$\rightarrow$FR, respectively. We generate instances that are labelled with one candidate and a binary value indicating whether it corresponds to the correct sense. we get 50,792 and 43,268 instances in DE$\rightarrow$EN and DE$\rightarrow$FR, respectively. 5K/5K examples are randomly selected as the test/development set. The remaining examples are used for training. We train 10 times with different seeds for each classifier and apply average accuracy. Table \ref{table-data-statis} lists the detailed statistics of the data. 
More experimental details are provided in the Appendix.

\begin{table}[htbp]
\begin{center}
\scalebox{0.92}{
\begin{tabular}{lrr}
\toprule
&DE$\rightarrow$EN &DE$\rightarrow$FR\\
\midrule
NMT training data&5.9M &2.1M\\
\addlinespace
Word senses&84&71\\
Lexical ambiguities&7,359&6,746\\
Instances & 50,792 &43,268\\
\bottomrule
\end{tabular}
}
\caption{\label{table-data-statis} Training data for NMT, and data extracted from ContraWSD: Word senses: total number of senses. Lexical ambiguities: number of sentences containing an ambiguous word. Instances: number of instances generated for WSD classification. }
\end{center}
\end{table}

\subsection{Results}
\label{sub:results}

Table \ref{table-result} provides the BLEU scores and the WSD accuracy on test sets, using different representations to represent ambiguous nouns. \textit{ENC} denotes encoder hidden states; \textit{DEC} means decoder hidden states. 

\begin{table}[htbp]
\begin{center}
\scalebox{0.92}{
\begin{tabular}{lcccc}
\toprule
& \multicolumn{2}{c}{DE$\rightarrow$EN}& \multicolumn{2}{c}{DE$\rightarrow$FR}\\
\cmidrule(l){2-3}
\cmidrule(l){4-5} &\textit{RNN.}&\textit{Trans.}&\textit{RNN.}&\textit{Trans.}\\
\midrule
BLEU & 29.1 & 32.6 & 17.0 & 19.3 \\
\addlinespace
\textit{Embedding} & 63.1 & 63.2 & 68.7 & 68.9 \\
\textit{ENC} & 94.2 & 97.2 & 91.7 & 95.6 \\
\textit{DEC} & 97.5 & 98.3 & 95.1 & 96.9 \\
\bottomrule
\end{tabular}
}
\caption{\label{table-result} BLEU scores of NMT models, and WSD accuracy on the test set using word embeddings or hidden states to represent ambiguous nouns. The hidden states are from the highest layer.\protect \footnotemark  ~\textit{RNN.}\  and \textit{Trans.}\  denote \textit{RNNS2S} and \textit{Transformer} models, respectively.}
\end{center}
\end{table}
\footnotetext{For encoders in \textit{RNNS2S}s, this is the last backward RNN.}

\noindent
\textit{ENC} achieves much higher accuracy than \textit{Embedding}. The WSD accuracy of \textit{Embedding} are around 63\% and 69\% in the two languages. While the accuracy of \textit{ENC} increases to over 91\%. The absolute accuracy gap varies from 23\% to 34\%, which is substantial. This result indicates that encoders have encoded a lot of relevant information for WSD into hidden states. 
In addition, \textit{DEC} achieves even higher accuracy than \textit{ENC} in both \textit{RNNS2S} models and \textit{Transformer} models.

\section{Analysis}
\label{analysis}

\subsection{WSD Classification} 
\label{WSD_classification}

\subsubsection{RNNS2S vs. Transformer} 
\textit{RNNS2S}s are inferior to \textit{Transformer}s distinctly in BLEU score. However, the hidden states from \textit{RNNS2S} also improve accuracy significantly, just not as much as those from \textit{Transformer} models. 
This result indicates that \textit{Transformer}s encode more relevant context for WSD than \textit{RNNS2S}s and accords with the finding in \newcite{Tang2018why} that \textit{Transformer}s perform WSD better than \textit{RNNS2S}s. 

The results of \textit{ENC} using \textit{RNNS2S} in Table \ref{table-result} are only based on hidden states from the last backward RNN. We also concatenate the hidden states from both forward and backward RNNs and get higher accuracy, 96.8\% in DE$\rightarrow$EN and 95.7\% in DE$\rightarrow$FR. The WSD accuracy of using bidirectional hidden states are competitive to using hidden states from Transformer models. However, concatenating forward and backward hidden states doubles the dimension. Thus, the comparison is not completely fair. 

\subsubsection{Encoder Depth}
\begin{figure}[htbp]
\centering
        \includegraphics[totalheight=4.1cm]{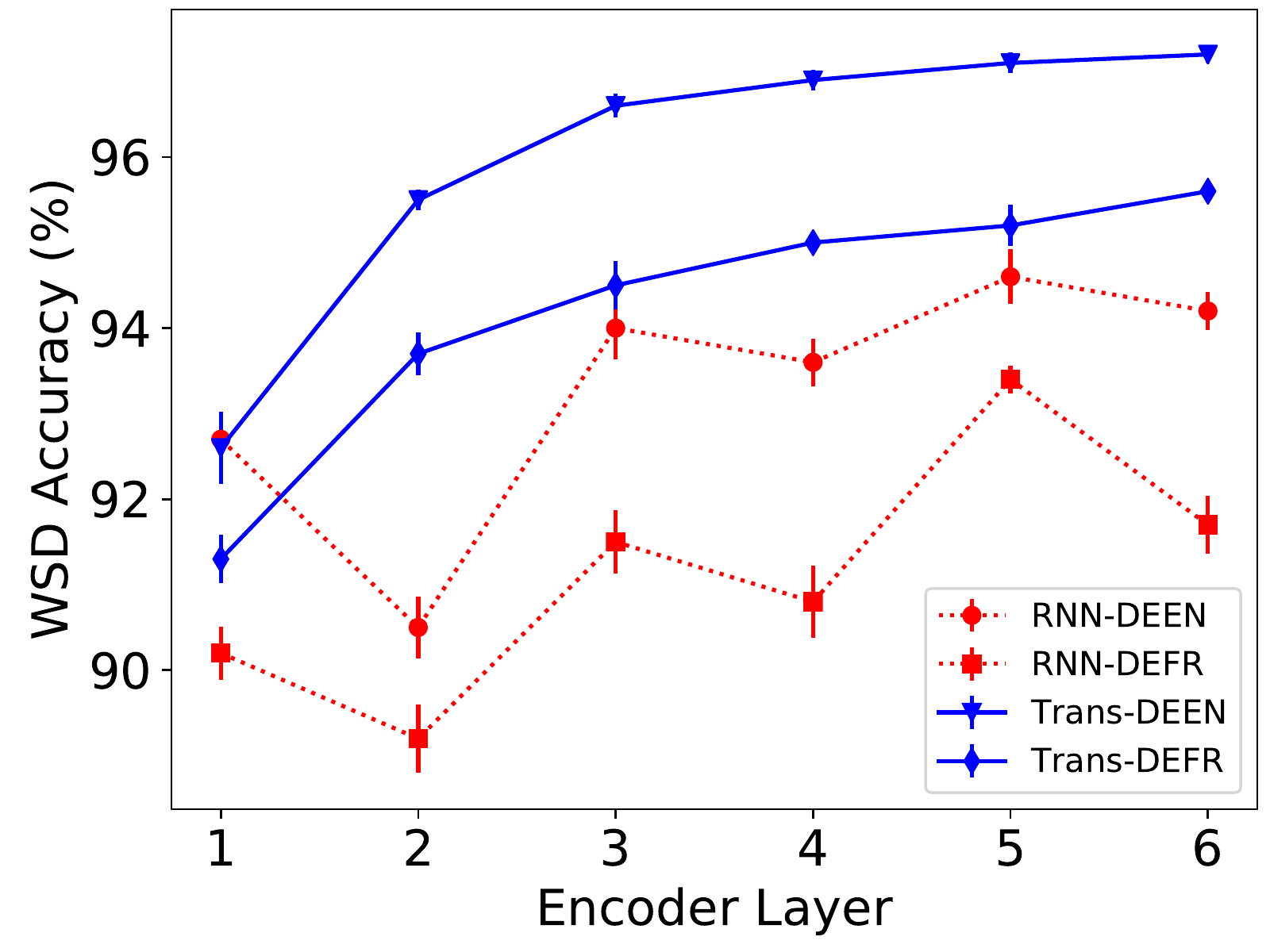}
    \caption{The WSD accuracy of using hidden states in different encoder layers, with standard deviation as error bars. For \textit{RNNS2S}s, the odd layers (1, 3, 5) are forward RNNs and the even layers (2, 4, 6) are backward RNNs. }
    \label{fig:acc-wsd}
\end{figure}

Figure \ref{fig:acc-wsd} illustrates WSD accuracy in different encoder layers, with standard deviation as error bars. Even the hidden states from the first layer boost the WSD performance substantially compared to using word embeddings. This means that most of the relevant information for WSD has been encoded into hidden states in the first encoder layer. 
For \textit{Transformer}s, the WSD accuracy goes up consistently as the encoder layer gets higher. 
\textit{RNNS2S} has 3 stacked bi-directional RNNs. Both forward and backward layers get higher accuracy when the depth increases. All the models show that hidden states in higher layers incorporate more relevant information for WSD. 

Our results conflict with the findings in \newcite{marvin2018exploring} where they find that hidden states in higher encoder layers do not perform disambiguation better than those in lower layers. One of the distinct differences from \newcite{marvin2018exploring} is that we train the classifier with $\sim$40K instances. While they employ 426 examples. Moreover, they extract encoder hidden states from NMT models with different layers rather than different layers of the same model. 

Moreover, it is interesting that the forward layers surpass the backward layers in the same bi-directional RNN. One possible explanation is that there is more relevant information for WSD before ambiguous nouns rather than after ambiguous nouns, which makes forward RNNs inject more relevant information into the hidden states of ambiguous nouns than backward RNNs. 

\subsubsection{Decoders}
\label{ssub:decoder}

As Table 2 shows, decoder hidden states could further improve the classification accuracy which accords with our hypothesis. It implies that the relevant information for WSD in the target-side has been well incorporated into the decoder hidden states to predict the translations of ambiguous nouns. 
      
Although decoder hidden states achieve higher accuracy than encoder hidden states, the improvement is not as big as that achieved by encoder hidden states over word embeddings. This indicates that most of the disambiguation work is done by encoders. 
\subsection{Self-attention}
\label{sub:self-attention}

\subsubsection{Attention Weights}
\label{ssub:attention_weights}

\begin{figure}[htbp]
\centering
        \includegraphics[totalheight=4.5cm]{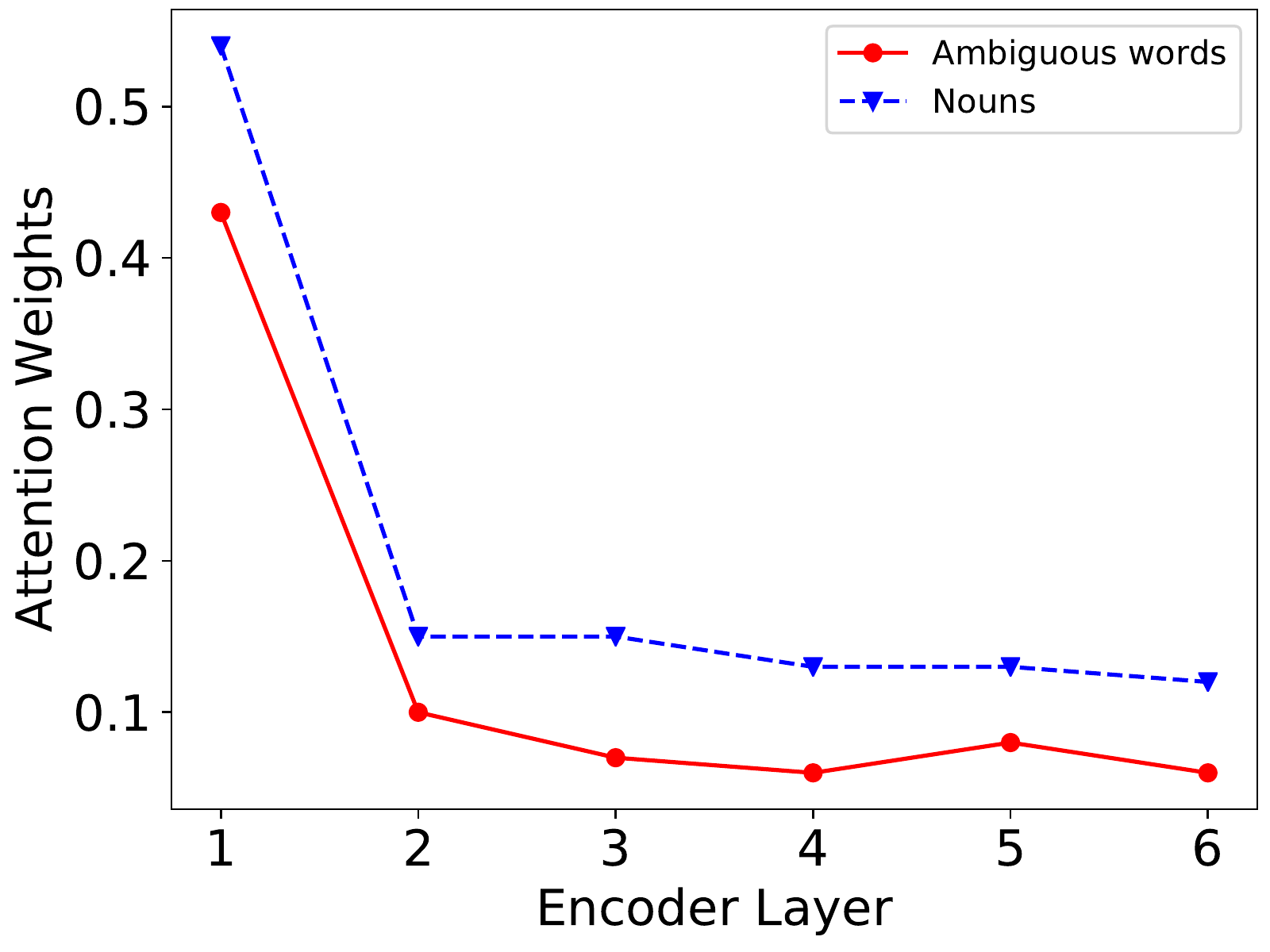}
    \caption{The average attention weights of ambiguous nouns and general nouns over themselves in different layers, in DE$\rightarrow$EN (same pattern in DE$\rightarrow$FR).}
    \label{fig:weights}     
\end{figure}

Figure \ref{fig:weights} exhibits the average attention weights of ambiguous nouns and all nouns over themselves in different layers. In the first layer, the attention weights are distinctly higher than those in higher layers. 87\% and 90\% of ambiguous nouns assign the highest attention to themselves in DE$\rightarrow$EN and DE$\rightarrow$FR, respectively. The attention weights drop dramatically from the second layer. It thus seems that self-attention pays more attention to the ambiguous nouns themselves in the first layer and 
to context words in the following layers. 

\noindent
The attention weights of ambiguous nouns are lower than those of nouns in general. That is, more attention is distributed to context words, which implies that self-attention recognizes ambiguous nouns and distributes more attention to the context.
We can conclude that self-attention pays more attention to context words to extract relevant information for disambiguation in all the layers, compared to nouns in general. 

\subsubsection{Attention Entropy}
\label{ssub:attention_entropy}

\begin{figure}[htbp]
\centering
        \includegraphics[totalheight=4.4cm]{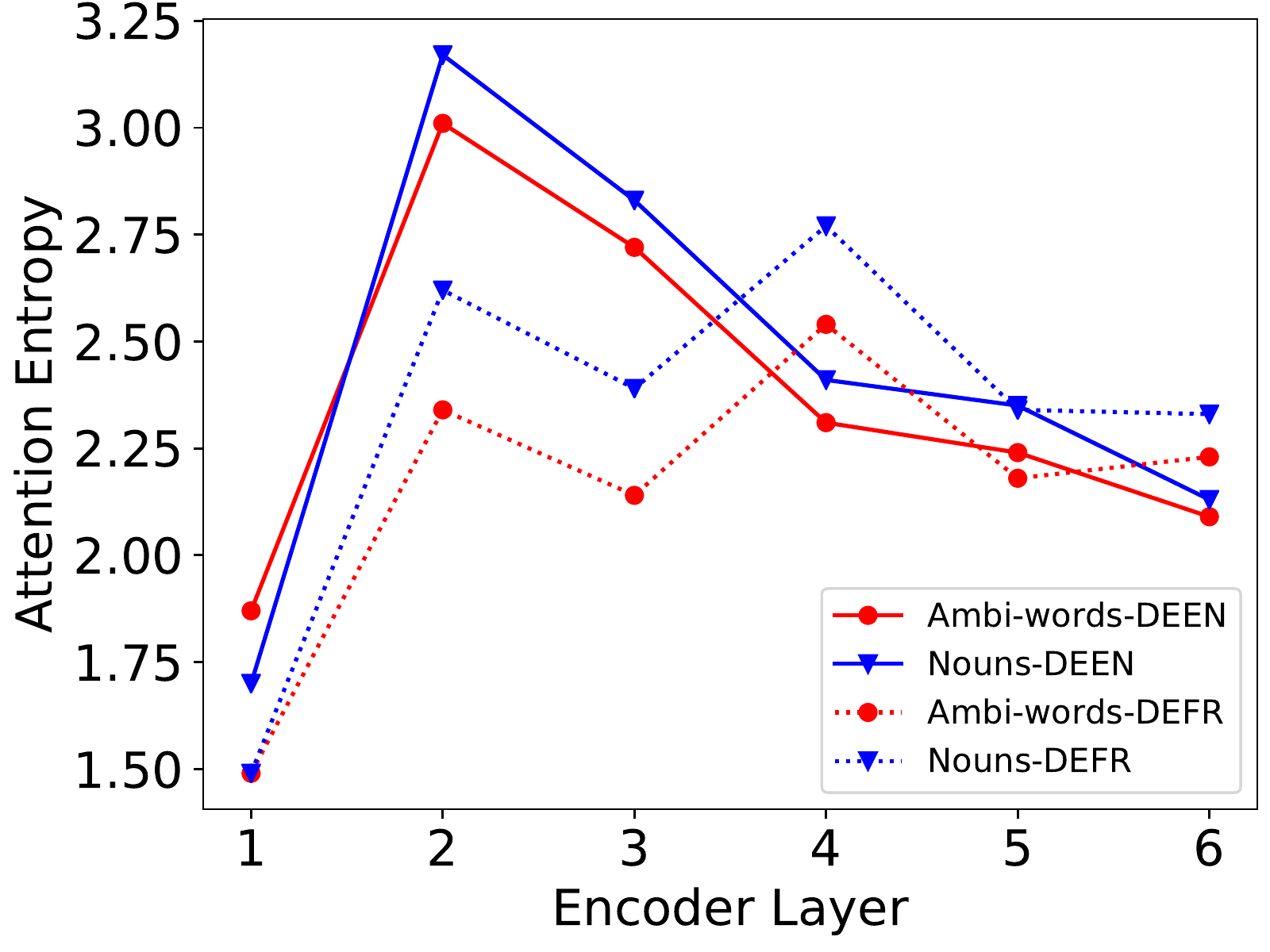}
    \caption{The average attention entropy of ambiguous nouns and nouns in different encoder layers.}
    \label{fig:entropy}
\end{figure}

Section \ref{ssub:attention_weights} has shown that self-attention of ambiguous nouns distributes more attention to the context than self-attention of nouns in general but what does the attention distribution look like? Figure \ref{fig:entropy} displays the average attention entropy of ambiguous nouns and all nouns in different layers. 
From the second layer, ambiguous nouns have smaller attention entropy than nouns in general, which means that self-attention mainly distributes attention to some specific words rather than all the words. As self-attention focuses on the ambiguous nouns themselves in the first layer, this result accords with our hypothesis as well. 

\noindent
In addition, there is a roughly general pattern that the attention entropy first rises and then drops. A plausible explanation is that the attention entropy first rises because context information is extracted from the entire sentence and later drops due to focusing on the most relevant context tokens. 

\section{Conclusion}

In this paper, we investigate the ability of NMT encoders and decoders to disambiguate word senses. 
We first train a neural classifier to predict whether the translation is correct given the representations of ambiguous nouns. We find that encoder hidden states outperform word embeddings significantly in the classification task which indicates that relevant information for WSD has been well integrated by encoders. In addition, the higher the encoder layer, the more relevant information is encoded into hidden states. Even though decoders could provide more relevant information for disambiguation, most of the disambiguation work is done by encoders.

We further explore the attention distributions of self-attention in encoders. The results show that self-attention can detect ambiguous nouns and distribute more attention to context words. 
Besides, self-attention focuses on the ambiguous nouns themselves in the first layer, then keeps extracting features from context words in higher layers.

\section*{Acknowledgments}
We thank all reviewers for their valuable and insightful comments. 
We acknowledge the computational resources provided by CSC in Helsinki and Sigma2 in Oslo through NeIC-NLPL (www.nlpl.eu). 
GT is mainly funded by the Chinese Scholarship Council (NO. 201607110016).

\bibliographystyle{acl_natbib}
\bibliography{eval-encoder}

\begin{thebibliography}{27}
\expandafter\ifx\csname natexlab\endcsname\relax\def\natexlab#1{#1}\fi

\bibitem[{Bahdanau et~al.(2015)Bahdanau, Cho, and Bengio}]{bahdanau15joint}
Dzmitry Bahdanau, Kyunghyun Cho, and Yoshua Bengio. 2015.
\newblock \href {https://arxiv.org/abs/1409.0473} {Neural machine translation
  by jointly learning to align and translate}.
\newblock In \emph{Proceedings of the 3rd International Conference on Learning
  Representations}, San Diego, California, USA.

\bibitem[{Belinkov et~al.(2017{\natexlab{a}})Belinkov, Durrani, Dalvi, Sajjad,
  and Glass}]{belinkov2017what}
Yonatan Belinkov, Nadir Durrani, Fahim Dalvi, Hassan Sajjad, and James Glass.
  2017{\natexlab{a}}.
\newblock \href {http://aclweb.org/anthology/P17-1080} {What do neural machine
  translation models learn about morphology?}
\newblock In \emph{Proceedings of the 55th Annual Meeting of the Association
  for Computational Linguistics (Volume 1: Long Papers)}, pages 861--872,
  Vancouver, Canada. Association for Computational Linguistics.

\bibitem[{Belinkov et~al.(2017{\natexlab{b}})Belinkov, M\`{a}rquez, Sajjad,
  Durrani, Dalvi, and Glass}]{belinkov2017evaluating}
Yonatan Belinkov, Llu\'{i}s M\`{a}rquez, Hassan Sajjad, Nadir Durrani, Fahim
  Dalvi, and James Glass. 2017{\natexlab{b}}.
\newblock \href {http://www.aclweb.org/anthology/I17-1001} {Evaluating layers
  of representation in neural machine translation on part-of-speech and
  semantic tagging tasks}.
\newblock In \emph{Proceedings of the Eighth International Joint Conference on
  Natural Language Processing (Volume 1: Long Papers)}, pages 1--10, Taipei,
  Taiwan. Asian Federation of Natural Language Processing.

\bibitem[{Bojar et~al.(2017)Bojar, Chatterjee, Federmann, Graham, Haddow,
  Huang, Huck, Koehn, Liu, Logacheva, Monz, Negri, Post, Rubino, Specia, and
  Turchi}]{wmt17}
Ond{\v{r}}ej Bojar, Rajen Chatterjee, Christian Federmann, Yvette Graham, Barry
  Haddow, Shujian Huang, Matthias Huck, Philipp Koehn, Qun Liu, Varvara
  Logacheva, Christof Monz, Matteo Negri, Matt Post, Raphael Rubino, Lucia
  Specia, and Marco Turchi. 2017.
\newblock \href {http://aclweb.org/anthology/W17-4717} {Findings of the 2017
  conference on machine translation ({WMT}17)}.
\newblock In \emph{Proceedings of the Second Conference on Machine
  Translation}, pages 169--214, Copenhagen, Denmark. Association for
  Computational Linguistics.

\bibitem[{Cho et~al.(2014)Cho, van Merrienboer, Gulcehre, Bahdanau, Bougares,
  Schwenk, and Bengio}]{cho2014learning}
Kyunghyun Cho, Bart van Merrienboer, Caglar Gulcehre, Dzmitry Bahdanau, Fethi
  Bougares, Holger Schwenk, and Yoshua Bengio. 2014.
\newblock \href {http://www.aclweb.org/anthology/D14-1179} {Learning phrase
  representations using {RNN} encoder--decoder for statistical machine
  translation}.
\newblock In \emph{Proceedings of the 2014 Conference on Empirical Methods in
  Natural Language Processing}, pages 1724--1734, Doha, Qatar. Association for
  Computational Linguistics.

\bibitem[{Ghader and Monz(2017)}]{ghader2017what}
Hamidreza Ghader and Christof Monz. 2017.
\newblock \href {http://www.aclweb.org/anthology/I17-1004} {What does attention
  in neural machine translation pay attention to?}
\newblock In \emph{Proceedings of the Eighth International Joint Conference on
  Natural Language Processing (Volume 1: Long Papers)}, pages 30--39, Taipei,
  Taiwan. Asian Federation of Natural Language Processing.

\bibitem[{Hieber et~al.(2017)Hieber, Domhan, Denkowski, Vilar, Sokolov,
  Clifton, and Post}]{Hieber2017sockeye}
Felix Hieber, Tobias Domhan, Michael Denkowski, David Vilar, Artem Sokolov, Ann
  Clifton, and Matt Post. 2017.
\newblock \href {http://arxiv.org/abs/1712.05690} {Sockeye: A toolkit for
  neural machine translation}.
\newblock \emph{arXiv preprint arXiv:1712.05690}.

\bibitem[{Kalchbrenner and Blunsom(2013)}]{kal2013recurrent}
Nal Kalchbrenner and Phil Blunsom. 2013.
\newblock \href {http://www.aclweb.org/anthology/D13-1176} {Recurrent
  continuous translation models}.
\newblock In \emph{Proceedings of the 2013 Conference on Empirical Methods in
  Natural Language Processing}, pages 1700--1709, Seattle, Washington, USA.
  Association for Computational Linguistics.

\bibitem[{Kingma and Ba(2015)}]{Kingma2014AdamAM}
Diederik~P. Kingma and Jimmy Ba. 2015.
\newblock \href {https://arxiv.org/abs/1412.6980} {Adam: A method for
  stochastic optimization}.
\newblock In \emph{Proceedings of the 3rd International Conference on Learning
  Representations}, San Diego, California, USA.

\bibitem[{Koehn(2005)}]{koehn2005epc}
Philipp Koehn. 2005.
\newblock {Europarl: A Parallel Corpus for Statistical Machine Translation}.
\newblock In \emph{{Proceedings of the 10th Machine Translation Summit}}, pages
  79--86, Phuket, Thailand.

\bibitem[{Koehn and Knowles(2017)}]{koehn2017challenges}
Philipp Koehn and Rebecca Knowles. 2017.
\newblock \href {http://www.aclweb.org/anthology/W17-3204} {Six challenges for
  neural machine translation}.
\newblock In \emph{Proceedings of the First Workshop on Neural Machine
  Translation}, pages 28--39, Vancouver, Canada. Association for Computational
  Linguistics.

\bibitem[{Luong et~al.(2015)Luong, Pham, and Manning}]{luong2015effective}
Thang Luong, Hieu Pham, and Christopher~D. Manning. 2015.
\newblock \href {http://aclweb.org/anthology/D15-1166} {Effective approaches to
  attention-based neural machine translation}.
\newblock In \emph{Proceedings of the 2015 Conference on Empirical Methods in
  Natural Language Processing}, pages 1412--1421, Lisbon, Portugal. Association
  for Computational Linguistics.

\bibitem[{Marvin and Koehn(2018)}]{marvin2018exploring}
Rebecca Marvin and Phillip Koehn. 2018.
\newblock \href
  {https://amtaweb.org/wp-content/uploads/2018/03/AMTA_2018_Proceedings_Research_Track.pdf#page=131}
  {Exploring word sense disambiguation abilities of neural machine translation
  systems}.
\newblock In \emph{Proceedings of AMTA 2018 (Volume 1: MT Research Track)},
  pages 125--131, Boston, USA. Association for Machine Translation in the
  Americas.

\bibitem[{Post(2018)}]{post2018sacre}
Matt Post. 2018.
\newblock \href {http://aclweb.org/anthology/W18-6319} {A call for clarity in
  reporting bleu scores}.
\newblock In \emph{Proceedings of the Third Conference on Machine Translation:
  Research Papers}, pages 186--191. Association for Computational Linguistics.

\bibitem[{Post and Vilar(2018)}]{Post2018fast}
Matt Post and David Vilar. 2018.
\newblock \href {https://www.aclweb.org/anthology/N18-1119} {Fast lexically
  constrained decoding with dynamic beam allocation for neural machine
  translation}.
\newblock In \emph{Proceedings of the 2018 Conference of the North {A}merican
  Chapter of the Association for Computational Linguistics: Human Language
  Technologies, Volume 1 (Long Papers)}, pages 1314--1324, New Orleans,
  Louisiana, USA. Association for Computational Linguistics.

\bibitem[{Rios et~al.(2017)Rios, Mascarell, and Sennrich}]{rios2017improving}
Annette Rios, Laura Mascarell, and Rico Sennrich. 2017.
\newblock \href {https://www.aclweb.org/anthology/W17-4702} {Improving word
  sense disambiguation in neural machine translation with sense embeddings}.
\newblock In \emph{Proceedings of the Second Conference on Machine
  Translation}, pages 11--19, Copenhagen, Denmark. Association for
  Computational Linguistics.

\bibitem[{Rios et~al.(2018)Rios, M{\"u}ller, and Sennrich}]{rios2018wsd}
Annette Rios, Mathias M{\"u}ller, and Rico Sennrich. 2018.
\newblock \href {https://www.aclweb.org/anthology/W18-6437} {The word sense
  disambiguation test suite at {WMT}18}.
\newblock In \emph{Proceedings of the Third Conference on Machine Translation:
  Shared Task Papers}, pages 588--596, Belgium, Brussels. Association for
  Computational Linguistics.

\bibitem[{Schmid(1995)}]{Schmid1995treetagger}
Helmut Schmid. 1995.
\newblock Improvements in part-of-speech tagging with an application to german.
\newblock In \emph{Proceedings of the ACL SIGDAT-Workshop}, Dublin, Ireland.
  Association for Computational Linguistics.

\bibitem[{Sennrich et~al.(2016)Sennrich, Haddow, and Birch}]{sennrich16sub}
Rico Sennrich, Barry Haddow, and Alexandra Birch. 2016.
\newblock \href {http://www.aclweb.org/anthology/P16-1162} {Neural machine
  translation of rare words with subword units}.
\newblock In \emph{Proceedings of the 54th Annual Meeting of the Association
  for Computational Linguistics (Volume 1: Long Papers)}, pages 1715--1725,
  Berlin, Germany. Association for Computational Linguistics.

\bibitem[{Sutskever et~al.(2014)Sutskever, Vinyals, and
  Le}]{sutskever2014sequence}
Ilya Sutskever, Oriol Vinyals, and Quoc~V Le. 2014.
\newblock \href
  {https://papers.nips.cc/paper/5346-sequence-to-sequence-learning-with-neural-networks.pdf}
  {Sequence to sequence learning with neural networks}.
\newblock In \emph{Proceedings of the Neural Information Processing Systems
  2014}, pages 3104--3112, Montr\'eal, Canada.

\bibitem[{Tang et~al.(2018{\natexlab{a}})Tang, M{\"u}ller, Rios, and
  Sennrich}]{Tang2018why}
Gongbo Tang, Mathias M{\"u}ller, Annette Rios, and Rico Sennrich.
  2018{\natexlab{a}}.
\newblock \href {http://aclweb.org/anthology/D18-1458} {Why self-attention? a
  targeted evaluation of neural machine translation architectures}.
\newblock In \emph{Proceedings of the 2018 Conference on Empirical Methods in
  Natural Language Processing}, pages 4263--4272, Brussels, Belgium.
  Association for Computational Linguistics.

\bibitem[{Tang et~al.(2018{\natexlab{b}})Tang, Sennrich, and
  Nivre}]{Tang2018WSD}
Gongbo Tang, Rico Sennrich, and Joakim Nivre. 2018{\natexlab{b}}.
\newblock \href {http://aclweb.org/anthology/W18-6304} {An analysis of
  attention mechanisms: The case of word sense disambiguation in neural machine
  translation}.
\newblock In \emph{Proceedings of the Third Conference on Machine Translation:
  Research Papers}, pages 26--35, Belgium, Brussels. Association for
  Computational Linguistics.

\bibitem[{Tang et~al.(2019{\natexlab{a}})Tang, Sennrich, and
  Nivre}]{tang-etal-2019-encoders}
Gongbo Tang, Rico Sennrich, and Joakim Nivre. 2019{\natexlab{a}}.
\newblock \href {https://doi.org/10.18653/v1/D19-1149} {Encoders help you
  disambiguate word senses in neural machine translation}.
\newblock In \emph{Proceedings of the 2019 Conference on Empirical Methods in
  Natural Language Processing and the 9th International Joint Conference on
  Natural Language Processing (EMNLP-IJCNLP)}, pages 1429--1435, Hong Kong,
  China. Association for Computational Linguistics.

\bibitem[{Tang et~al.(2019{\natexlab{b}})Tang, Sennrich, and
  Nivre}]{tang2019understanding}
Gongbo Tang, Rico Sennrich, and Joakim Nivre. 2019{\natexlab{b}}.
\newblock Understanding neural machine translation by simplification: The case
  of encoder-free models.
\newblock In \emph{Proceedings of the International Conference Recent Advances
  in Natural Language Processing, {RANLP} 2019}, Varna, Bulgaria.

\bibitem[{Vaswani et~al.(2017)Vaswani, Shazeer, Parmar, Uszkoreit, Jones,
  Gomez, Kaiser, and Polosukhin}]{vaswani2017Attention}
Ashish Vaswani, Noam Shazeer, Niki Parmar, Jakob Uszkoreit, Llion Jones,
  Aidan~N Gomez, \L~ukasz Kaiser, and Illia Polosukhin. 2017.
\newblock \href
  {http://papers.nips.cc/paper/7181-attention-is-all-you-need.pdf} {Attention
  is all you need}.
\newblock In \emph{Advances in Neural Information Processing Systems 30}, pages
  6000--6010. Curran Associates, Inc.

\bibitem[{Voita et~al.(2018)Voita, Serdyukov, Sennrich, and
  Titov}]{voita2018context}
Elena Voita, Pavel Serdyukov, Rico Sennrich, and Ivan Titov. 2018.
\newblock \href {https://www.aclweb.org/anthology/P18-1117} {Context-aware
  neural machine translation learns anaphora resolution}.
\newblock In \emph{Proceedings of the 56th Annual Meeting of the Association
  for Computational Linguistics (Volume 1: Long Papers)}, pages 1264--1274,
  Melbourne, Australia. Association for Computational Linguistics.

\bibitem[{Voita et~al.(2019)Voita, Talbot, Moiseev, Sennrich, and
  Titov}]{voita-etal-2019-analyzing}
Elena Voita, David Talbot, Fedor Moiseev, Rico Sennrich, and Ivan Titov. 2019.
\newblock \href {https://www.aclweb.org/anthology/P19-1580} {Analyzing
  multi-head self-attention: Specialized heads do the heavy lifting, the rest
  can be pruned}.
\newblock In \emph{Proceedings of the 57th Annual Meeting of the Association
  for Computational Linguistics}, pages 5797--5808, Florence, Italy.
  Association for Computational Linguistics.

\end{thebibliography}

\clearpage
\appendix
\section{Appendix}
\label{sec:appendix}
\subsection{Data} 
\label{ssec:data}

\paragraph{NMT}
For DE$\rightarrow$EN, we use \textit{newstest2013} as the validation set, and use \textit{newstest2017} as the test set. For DE$\rightarrow$FR, we use \textit{newstest2013} as the evaluation set, and use \textit{newstest2012} as the test set.

\subsection{Experimental Settings}
\label{ssec:settings}

\paragraph{NMT}
We implemented \textit{RNNS2S} models with stacked bi-directional RNNs and implemented the self-attention in Transformer encoders to output the attention distributions.
We use \textit{Adam} \cite{Kingma2014AdamAM} as the optimizer. 
The initial learning rate is set to 0.0002. 
All the neural networks have 6 layers.\footnote{The RNN encoder is a stack of three bi-directional RNNs which is equivalent to 6 uni-directional RNNs.} 
The size of embeddings and hidden units is 512.
The attention mechanism in \textit{Transformer} has 8 heads. 
We learn a joint BPE model with 32,000 subword units \cite{sennrich16sub}. 
All BLEU scores are computed with \textit{SacreBLEU} \cite{post2018sacre}.

\paragraph{WSD Classification}
The classifiers are feed-forward neural networks with only one hidden layer, using ReLU non-linear activation. The size of the hidden layer is set to 512. We use Adam learning algorithm as well with mini-batches of size 3,000. 
The classifiers are trained using a cross-entropy loss. Each classifier is trained for 80 epochs and the one performs best on the development set is selected for evaluation. 

\end{document}